\documentclass[11pt,a4paper]{article}
\usepackage[hyperref]{emnlp2018}
\usepackage{times}
\usepackage{latexsym}

\usepackage{url}
\usepackage{graphicx}
\usepackage{amsmath}
\usepackage{amssymb}
\DeclareMathOperator*{\argmax}{argmax}

\aclfinalcopy 

\title{Simple Fusion: Return of the Language Model}

\author{Felix Stahlberg$^\dag$\footnotemark \and James Cross$^{\ddagger}$ \and Veselin Stoyanov$^{\ddagger}$\\
 $^\dag$Department of Engineering, University of Cambridge, UK  \\
 $^\ddagger$Applied Machine Learning, Facebook, Menlo Park, CA, USA \\
  {\tt fs439@cam.ac.uk}, {\tt jcross@fb.com}, {\tt vesko.st@gmail.com} \\
  }

\date{}

\begin{document}
\maketitle
\begin{abstract}
Neural Machine Translation (NMT) typically leverages monolingual data in training through backtranslation. We investigate an alternative simple method to use monolingual data for NMT training: We combine the scores of a pre-trained and fixed language model (LM) with the scores of a translation model (TM) while the TM is trained from scratch. To achieve that, we train the translation model to predict the residual probability of the training data added to the prediction of the LM. This enables the TM to focus its capacity on modeling the source sentence since it can rely on the LM for fluency. We show that our method outperforms previous approaches to integrate LMs into NMT while the architecture is simpler as it does not require gating networks to balance TM and LM. We observe gains of between +0.24 and +2.36 BLEU on all four test sets (English-Turkish, Turkish-English, Estonian-English, Xhosa-English) on top of ensembles without LM. We compare our method with alternative ways to utilize monolingual data such as backtranslation, shallow fusion, and cold fusion.
\end{abstract}

\footnotetext{This work was done when the first author was on an internship at Facebook.}

\section{Introduction}

Machine translation (MT) relies on parallel training data, which is difficult to acquire. In contrast, monolingual data is abundant for most languages and domains. Traditional statistical machine translation (SMT) effectively leverages monolingual data using language models (LMs)~\citep{lm-in-mt}. The combination of LM and TM in SMT can be traced back to the noisy-channel model which applies the Bayes rule to decompose a translation system~\citep{ibm}:

\begin{equation}
\begin{split}
\hat{\mathbf{y}} =& \argmax_\mathbf{y} P(\mathbf{y}|\mathbf{x}) \\
=& \argmax_\mathbf{y} P_{TM}(\mathbf{x}|\mathbf{y})P_{LM}(\mathbf{y})
\end{split}
\label{eq:noisy-channel}
\end{equation}
where $\mathbf{x}=(x_1,\dots,x_m)$ is the source sentence, $\mathbf{y}=(y_1,\dots,y_n)$ is the target sentence, and $P_{TM}(\cdot)$ and $P_{LM}(\cdot)$ are translation model and language model probabilities.

In contrast, NMT~\citep{sutskever,bahdanau} uses a discriminative model and learns the distribution  $P(\mathbf{y}|\mathbf{x})$ directly end-to-end. Therefore, the vanilla training regimen for NMT is not amenable to integrating an LM or monoglingual data in a straightforward manner.

An early attempt to use LMs for NMT, also known as {\em shallow fusion}, combines LM and NMT scores at inference time in a log-linear model~\citep{deepfusion,deepfusion-csl}. In contrast, we integrate the LM scores during NMT training. Our training procedure first trains an LM on a large monolingual corpus. We then hold the LM fixed and train the NMT system to optimize the combined score of LM and NMT on the parallel training set. This allows the NMT model to focus on modeling the source sentence, while the LM handles the generation based on the target-side history. \citet{coldfusion} explored a similar idea for speech recognition using a gating network for controlling the relative contribution of the LM. We show that our simpler architecture without an explicit control mechanism is effective for machine translation. We observe gains of up to more than 2 BLEU points from adding the LM to TM training. We also show that our method can be combined with backtranslation~\citep{backtranslation}, yielding further gains over systems without LM. 

\section{Related Work}

\subsection{Inference-time Combination}
\label{sec:shallow-fusion}

{\em Shallow fusion}~\citep{deepfusion} integrates an LM by changing the decoding objective to:

\begin{equation}
\hat{\mathbf{y}} = \argmax_\mathbf{y} \log P_\text{TM}(\mathbf{y}|\mathbf{x}) + \lambda\log P_\text{LM}(\mathbf{y}).
\label{eq:shallow-fusion}
\end{equation}
$P_\text{LM}(\cdot)$ is produced by an LSTM-based RNN-LM~\citep{rnnlm} which has been trained on monolingual target language data. $P_\text{TM}(\cdot)$ can be a typical encoder-decoder Seq2Seq model~\citep{sutskever,bahdanau,dot-attn}. $\lambda$ is a hyper-parameter which is tuned on the development set.

\subsection{Cold Fusion}
\label{sec:cold-fusion}

Shallow fusion combines a fixed TM with a fixed LM at inference time. \citet{coldfusion} proposed to keep the LM fixed, but train a sequence to sequence (Seq2Seq) NMT model from scratch which includes the LM as a fixed part of the network. They argue that this approach allows the Seq2Seq network to use its model capacity for the conditioning on the source sequence since the language modeling aspect is already covered by the LM. Their {\em cold fusion} architecture includes a gating network which learns to regulate the contributions of the LM at each time step. They demonstrated superior performance of cold fusion on a speech recognition task.

\subsection{Other Approaches}

\citet{deepfusion,deepfusion-csl} suggest to combine a pre-trained RNN-LM with a pre-trained NMT system using a controller network that dynamically adjusts the weights between RNN-LM and NMT at each time step ({\em deep fusion}). Both deep fusion and $n$-best reranking with count-based LMs have been used in WMT evaluation systems~\citep{wmt15-montreal,wmt17-sogou}. An important limitation of these approaches is that LM and TM are trained independently. 

A second line of research augments the parallel training data with additional synthetic data from a monolingual corpus in the target language. The source sentences can be generated with a separate translation system~\citep{backtranslation-smt,backtranslation} (‘backtranslation’), or simply copied over from the target side~\citep{copytarget}. Since data augmentation methods rely on some balance between real and synthetic data~\citep{backtranslation,copytarget,investigating-backtranslation}, they can often only use a small fraction of the available monolingual data. A third class of approaches change the NMT training loss function to incorporate monolingual data. For example, \citet{autoencoder,reconstruction} proposed to add autoencoder terms to the training objective which capture how well a sentence can be reconstructed from its translated representation. However, training with respect to the new loss is often computationally intensive and requires approximations. Alternatively, multi-task learning has been used to incorporate source-side~\citep{source-multi-task} and target-side~\citep{target-multi-task} monolingual data. Another way of utilizing monolingual data in both source and target language is to warm start Seq2Seq training from pre-trained encoder and decoder networks~\citep{pretrain,pretrain-and-shallow-fusion}. We note that pre-training can be used in combination with our approach.

An extreme form of leveraging monolingual training data is unsupervised NMT~\citep{unsupervised,unsupervised2} which removes the need for parallel training data entirely. In this work, we assume to have access to some amount of parallel training data, but aim to improve the translation quality even further by using a language model.

\section{Translation Model Training under Language Model Predictions}

In spirit of the cold fusion technique of \citet{coldfusion} we also keep the LM fixed when training the translation network. However, we greatly simplify the architecture by removing the need for a gating network. We follow the usual left-to-right factorization in NMT:

\begin{equation}
 P(\mathbf{y}|\mathbf{x}) = \prod_{t=1}^n  P(y_t|y_1^{t-1},\mathbf{x}).
\end{equation}
Let $S_\text{TM}(y_t|y_1^{t-1},\mathbf{x})$ be the output of the TM projection layer without softmax, i.e.,\ what we would normally call the logits. We investigate two different ways to parameterize $P(y_t|y_1^{t-1},\mathbf{x})$ using $S_\text{TM}(y_t|y_1^{t-1},\mathbf{x})$ and a fixed and pre-trained language model $P_\text{LM}(\cdot)$: {\sc PostNorm} and {\sc PreNorm}.

\paragraph{{\sc PostNorm}}
This variant is directly inspired by shallow fusion (Eq.~\ref{eq:shallow-fusion}) as we turn $S_\text{TM}(y_t|y_1^{t-1},\mathbf{x})$ into a probability distribution using a softmax layer, and sum its log-probabilities with the log-probabilities of the LM, i.e.\ multiply their probabilities:

\begin{equation}
\begin{split}
P(y_t|y_1^{t-1},\mathbf{x}) =& \text{softmax}\big(\text{softmax}( \\
&S_\text{TM}(y_t|y_1^{t-1},\mathbf{x}))P_\text{LM}(y_t|y_1^{t-1})\big).
\end{split}
\label{eq:logprob}
\end{equation}

\paragraph{{\sc PreNorm}} 
Another option is to apply normalization after combining the raw $S_\text{TM}(y_t|y_1^{t-1},\mathbf{x})$ scores with the LM log-probability:
\begin{equation}
\begin{split}
 P(y_t|y_1^{t-1},\mathbf{x}) =& \text{softmax}\Big(S_\text{TM}(y_t|y_1^{t-1},\mathbf{x}) \\
&+ \log P_\text{LM}(y_t|y_1^{t-1}) \Big).
\end{split}
\label{eq:plain}
\end{equation}

\subsection{Theoretical Discussion of {\sc PostNorm} and {\sc PreNorm}}

The double softmax in {\sc PostNorm} is needed because the inner term (component-wise product of two distributions) would not be a valid probability distribution as it is not guaranteed to sum to 1. Another way to fix this issue would be to combine TM and LM probabilities in the probability space rather than in the log space. However, we have found that probability space combination does not work as well as {\sc PostNorm} in our experiments. We can describe $S_\text{TM}(y_t|y_1^{t-1},\mathbf{x})$ under {\sc PostNorm} informally as the residual probability added to the prediction of the LM.

It is interesting to investigate what signal is actually propagated into $S_\text{TM}(y_t|y_1^{t-1},\mathbf{x})$ when training with the {\sc PreNorm} strategy. We can rewrite $P(y_t|y_1^{t-1},\mathbf{x})$ as:

\begin{equation}
\begin{split}
P(y_t|y_1^{t-1},\mathbf{x}) =& \frac{P(y_t,y_1^{t-1}|\mathbf{x})}{P(y_1^{t-1}|\mathbf{x})} \\
=& \frac{P(y_t,\mathbf{x}|y_1^{t-1})}{P(\mathbf{x}|y_1^{t-1})} \\
=& \frac{P(\mathbf{x}|y_t, y_1^{t-1})}{P(\mathbf{x}|y_1^{t-1})} P(y_t|y_1^{t-1}).
\end{split}
\label{eq:plain-meaning}
\end{equation}
Alternatively, we can decompose $P(y_t|y_1^{t-1},\mathbf{x})$ as follows using Eq.~\ref{eq:plain}:
\begin{equation}
\begin{split}
P(y_t|y_1^{t-1},\mathbf{x}) =& \text{softmax}\Big(S_\text{TM}(y_t|y_1^{t-1},\mathbf{x}) \\
&+ \log P_\text{LM}(y_t|y_1^{t-1}) \Big) \\
\propto & \exp\Big(S_\text{TM}(y_t|y_1^{t-1},\mathbf{x}) \\
&+ \log P_\text{LM}(y_t|y_1^{t-1}) \Big) \\
=& \exp(S_\text{TM}(y_t|y_1^{t-1},\mathbf{x}))\\ 
&\cdot P_\text{LM}(y_t|y_1^{t-1}).
\end{split}
\label{eq:plain-meaning2}
\end{equation}
Combining Eq.~\ref{eq:plain-meaning} and Eq.~\ref{eq:plain-meaning2} leads to:
\begin{equation}
\exp(S_\text{TM}(y_t|y_1^{t-1},\mathbf{x}))\propto\frac{P(\mathbf{x}|y_1^t)}{P(\mathbf{x}|y_1^{t-1})}
\end{equation}
This means that $S_\text{TM}(y_t|y_1^{t-1},\mathbf{x})$ under {\sc PreNorm} is trained to predict how much more likely the {\em source} sentence becomes when a particular target token $y_t$ is revealed.

\section{Experimental Setup}

\begin{table}[t!]
\centering
\small
\begin{tabular}{|l|r|}
\hline
 {\bf Language pair} & {\bf \# Sentences} \\\hline
 Turkish-English (\textsc{WMT}) & 207.7K \\
 Estonian-English (\textsc{WMT}) & 2,178.0K \\
 Xhosa-English (\textsc{Internal}) & 739.2K \\
\hline
\end{tabular}
\caption{Parallel training data.}\label{tab:parallel}
\end{table}

\begin{table}[t!]
\centering
\small
\begin{tabular}{|l|r|r|r|}
\hline
 {\bf Language} & {\bf \# Sentences} & \multicolumn{2}{c|}{{\bf LM Perplexity}} \\
  &  & {\bf dev} & {\bf test} \\\hline
English (\textsc{WMT}) & 26.9M & 91.16 & 87.77  \\
Turkish (\textsc{WMT}) & 3.0M & 59.19 & 70.46  \\
English (\textsc{Internal}) & 20.0M & 105.28 & 108.19 \\
\hline
\end{tabular}
\caption{Monolingual training data.}\label{tab:lm}
\end{table}

We evaluate our method on a variety of publicly available and proprietary data sets. For our Turkish-English (tr-en), English-Turkish (en-tr), and Estonian-English (et-en) experiments we use all available parallel data from the WMT18 evaluation campaign to train the translation models. Our language models are trained on {\em News Crawl 2017}. 
We use {\em news-test2017} as development (``dev'') set and {\em news-test2018} as test set.

Additionally, we collected our own proprietary corpus of public posts on Facebook. We refer to it as `\textsc{Internal}' data set. This corpus consists of monolingual English in-domain sentences and parallel data in Xhosa-English. Training set sizes are summarized in Tables ~\ref{tab:parallel} and~\ref{tab:lm}.

Our preprocessing consists of lower-casing, tokenization, and subword-segmentation using joint byte pair encoding~\citep{bpe} with 16K merge operations. On Turkish, we additionally remove diacritics from the text.

On WMT we use lower-cased SacreBLEU\footnote{SacreBLEU signature for tr-en test-2017: \\
BLEU+c.lc+l.tr-en+\#.1+s.exp+t.wmt17+tok.13a+v.1.2.10}~\citep{sacrebleu} to be comparable with the literature.\footnote{For translation into Turkish we evaluate after diacritics removal.} On our internal data we report tokenized BLEU scores.

Our Seq2Seq models are encoder-decoder architectures~\citep{sutskever,bahdanau} with dot-product attention~\citep{luong-attention} trained with our PyTorch Translate library.\footnote{\url{https://github.com/pytorch/translate}} Both decoder and encoder consist of two 512-dimensional LSTM layers and 256-dimensional embeddings. The first encoder layer is bidirectional, the second one runs from right to left. Our training and architecture hyperparameters are summarized in Tab.~\ref{tab:hyperparam}. Our LSTM-based LMs have the same size and architecture as the decoder networks, but do not use attention and do not condition on the source sentence. We run beam search with beam size of 6 in all our experiments.

For each setup we train five models using SGD (batch size of 32 sentences) with learning rate decay and label smoothing, and either select the best one (single system) or ensemble the four best models based on dev set BLEU score.

\begin{table}[t!]
\centering
\small
\begin{tabular}{|l|r|}
\hline
\multicolumn{2}{|c|}{{\bf Architecture Hyperparameters}} \\
\hline
Source vocab size (BPE) & 16,000 \\
Target vocab size (BPE) & 16,000 \\
Embedding size (all) & 256 \\
Encoder LSTM units & 512 \\
Encoder layers & 2 \\
Decoder LSTM units & 512 \\
Decoder layers & 2 \\
Attention type & dot product \\
\hline \hline 
\multicolumn{2}{|c|}{{\bf Training Settings}} \\
\hline
Optimization & Vanilla SGD \\
Learning rate & 0.5 \\
Batch size & 32 \\
Label smoothing $\epsilon$ & 0.1 \\
Checkpoint averaging & Last 10 \\
\hline
\end{tabular}
\caption{Summary of NMT settings for all models.}\label{tab:hyperparam}
\end{table}

\section{Results}

\begin{table}[t!]
\centering
\small

\begin{tabular}{|l@{\hspace{0.2em}}|@{\hspace{0.4em}}c@{\hspace{0.4em}}|@{\hspace{0.4em}}c@{\hspace{0.4em}}|@{\hspace{0.4em}}c@{\hspace{0.4em}}|@{\hspace{0.4em}}c@{\hspace{0.4em}}|}
\multicolumn{5}{c}{{\bf English-Turkish (WMT)}} \\\hline
{\bf Method} & \multicolumn{2}{c}{{\bf Single}} & \multicolumn{2}{c|}{{\bf 4-Ensemble}} \\
 & {\bf dev} & {\bf test} & {\bf dev} & {\bf test} \\\hline
Baseline (no LM) & 12.23 & 11.56 & 14.17 & 13.35 \\
Shallow fusion & 12.45 & 11.61 & 14.43 & 13.51 \\
Cold fusion & 12.39 & 11.54 & 14.20 & 13.23 \\\hline
{\bf This work}: {\sc PreNorm} & 12.82 & 11.93 & {\bf 14.78} & 13.41 \\
{\bf This work}: {\sc PostNorm} & {\bf 13.30} & {\bf 12.27} & 14.77 & {\bf 13.61} \\
\hline
\end{tabular}

\vspace*{10pt}

\begin{tabular}{|l@{\hspace{0.2em}}|@{\hspace{0.4em}}c@{\hspace{0.4em}}|@{\hspace{0.4em}}c@{\hspace{0.4em}}|@{\hspace{0.4em}}c@{\hspace{0.4em}}|@{\hspace{0.4em}}c@{\hspace{0.4em}}|}
\multicolumn{5}{c}{{\bf Turkish-English (WMT)}} \\\hline
{\bf Method} & \multicolumn{2}{c}{{\bf Single}} & \multicolumn{2}{c|}{{\bf 4-Ensemble}} \\
 & {\bf dev} & {\bf test} & {\bf dev} & {\bf test} \\\hline
Baseline (no LM) & 16.14 & 16.60 & 18.01 & 18.67 \\
Shallow fusion & 16.11 & 16.70 & 18.01 & 18.67 \\
Cold fusion & 16.25 & 16.21 & 17.99 & 18.40 \\\hline
{\bf This work}: {\sc PreNorm} & 15.88 & 16.39 & 17.95 & 18.40 \\
{\bf This work}: {\sc PostNorm} & {\bf 16.59} & {\bf 17.03} & {\bf 18.38} & {\bf 19.17} \\
\hline
\end{tabular}

\vspace*{10pt}

\begin{tabular}{|l@{\hspace{0.2em}}|@{\hspace{0.4em}}c@{\hspace{0.4em}}|@{\hspace{0.4em}}c@{\hspace{0.4em}}|@{\hspace{0.4em}}c@{\hspace{0.4em}}|@{\hspace{0.4em}}c@{\hspace{0.4em}}|}
\multicolumn{5}{c}{{\bf Estonian-English (WMT)}} \\\hline
{\bf Method} & \multicolumn{2}{c}{{\bf Single}} & \multicolumn{2}{c|}{{\bf 4-Ensemble}} \\
 & {\bf dev} & {\bf test} & {\bf dev} & {\bf test} \\\hline
Baseline (no LM) & 16.02 & 16.57 & 16.83 & 17.91 \\
Shallow fusion & 16.02 & 16.57 & 16.83 & 17.91 \\
Cold fusion & 15.40 & 15.99 & 16.48 & 17.79 \\\hline
{\bf This work}: {\sc PreNorm} & {\bf 16.80} & {\bf 17.44} & {\bf 17.78} & {\bf 19.01} \\
{\bf This work}: {\sc PostNorm} & 16.43 & 17.10 & 17.62 & 18.63 \\
\hline
\end{tabular}

\vspace*{10pt}

\begin{tabular}{|l@{\hspace{0.2em}}|@{\hspace{0.4em}}c@{\hspace{0.4em}}|@{\hspace{0.4em}}c@{\hspace{0.4em}}|@{\hspace{0.4em}}c@{\hspace{0.4em}}|@{\hspace{0.4em}}c@{\hspace{0.4em}}|}
\multicolumn{5}{c}{{\bf Xhosa-English (\textsc{Internal)}}} \\\hline
{\bf Method} & \multicolumn{2}{c}{{\bf Single}} & \multicolumn{2}{c|}{{\bf 4-Ensemble}} \\
 & {\bf dev} & {\bf test} & {\bf dev} & {\bf test} \\\hline
Baseline (no LM) & 10.39 & 11.49 & 13.87 & 15.43 \\
Shallow fusion & 10.69 & 11.65 & 14.06 & 15.54 \\
Cold fusion & 10.72 & 11.29 & 13.66 & 15.13 \\\hline
{\bf This work}: {\sc PreNorm} & 11.06 & 12.13 & 14.50 & 16.07 \\
{\bf This work}: {\sc PostNorm} & {\bf 12.34} & {\bf 13.27} & {\bf 15.45} & {\bf 17.79} \\
\hline
\end{tabular}

\caption{Comparison of our {\sc PreNorm} and {\sc PostNorm} combination strategies with shallow fusion~\citep{deepfusion} and cold fusion~\citep{coldfusion} under an RNN-LM.}\label{tab:wmt-results}
\end{table}

Tab.~\ref{tab:wmt-results} compares our methods {\sc PreNorm} and {\sc PostNorm} on the tested language pairs. Shallow fusion (Sec.~\ref{sec:shallow-fusion}) often leads to minor improvements over the baseline for both single systems and ensembles. We also reimplemented the {\em cold fusion} technique (Sec.~\ref{sec:cold-fusion}) for comparison. For our machine translation experiments we report mixed results with cold fusion, with performance ranging between 0.33 BLEU gain on Xhosa-English and slight BLEU degradation in most of our Turkish-English experiments. 

Both of our methods, {\sc PreNorm} and {\sc PostNorm} yield significant improvements in BLEU across the board. We report more consistent gains with {\sc PostNorm} than with {\sc PreNorm}. All our {\sc PostNorm} systems outperform both shallow fusion and cold fusion on all language pairs, yielding test set gains of up to +2.36 BLEU (Xhosa-English ensembles).

\section{Discussion and Analysis}

\paragraph{Backtranslation}

\begin{figure}[t!]
\centering
\includegraphics[width=0.48\textwidth]{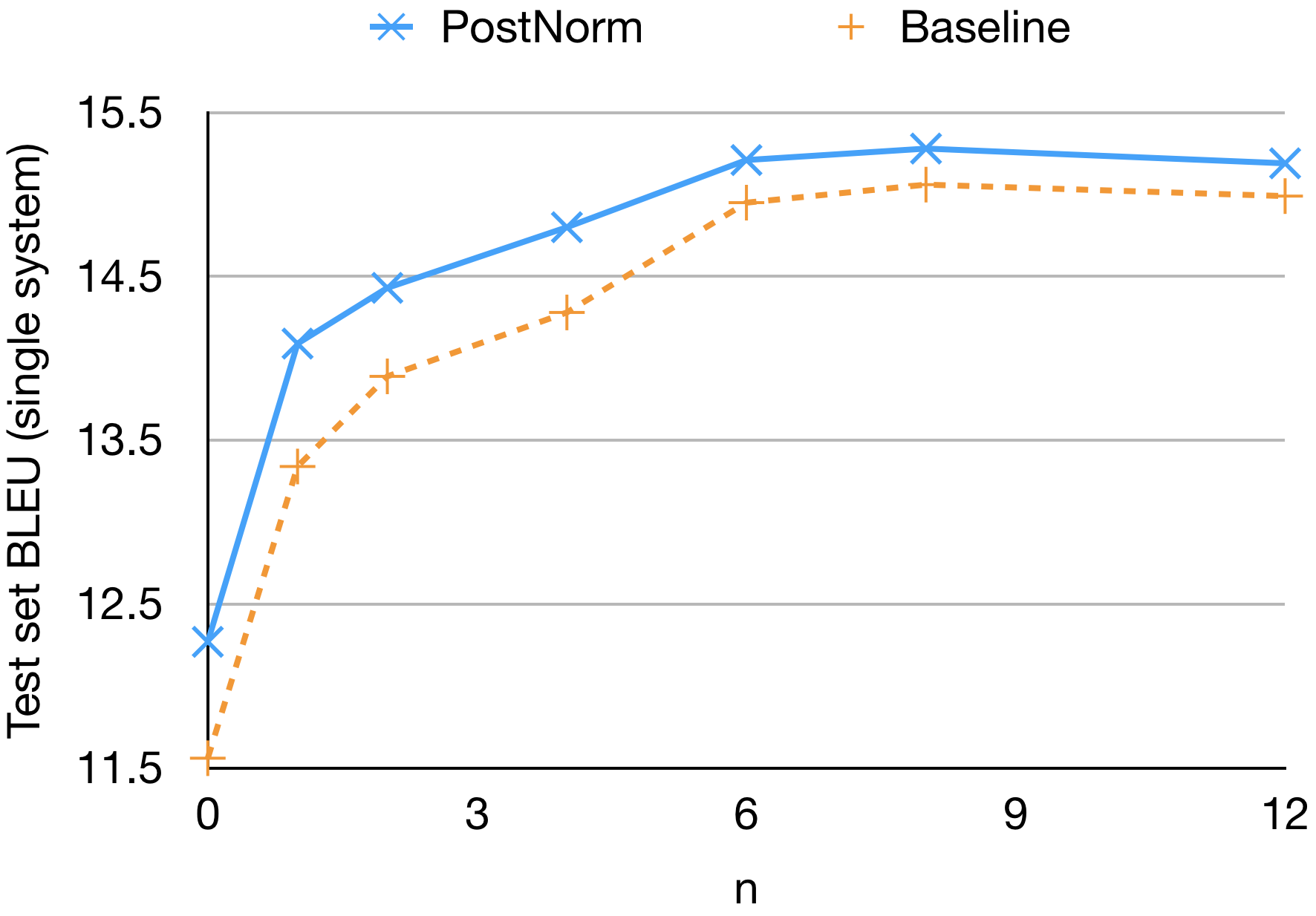}
\caption{Performance using backtranslation on English-Turkish. Synthetic sentences are mixed at a ratio of 1:$n$ where $n$ is plotted on the $x$-axis.}\label{fig:backtranslation}
\end{figure}

A very popular technique to use monolingual data for NMT is backtranslation~\citep{backtranslation}. Backtranslation uses a reverse NMT system to translate monolingual target language sentences into the source language, and adds the newly generated sentence pairs to the training data. The amount of monolingual data which can be used for backtranslation is usually limited by the size of the parallel corpus as the translation quality suffers when the mixing ratio between synthetic and real source sentences is too large~\citep{investigating-backtranslation}. This is a severe limitation particularly for low-resource MT. Fig.~\ref{fig:backtranslation} shows that both our baseline system without LM and our {\sc PostNorm} system benefit greatly from backtranslation up to a mixing ratio of 1:8, but degrade slightly if this ratio is exceeded. {\sc PostNorm} is significantly better than the baseline even when using it in combination with backtranslation.

\begin{figure}[t!]
\centering
\includegraphics[width=0.48\textwidth]{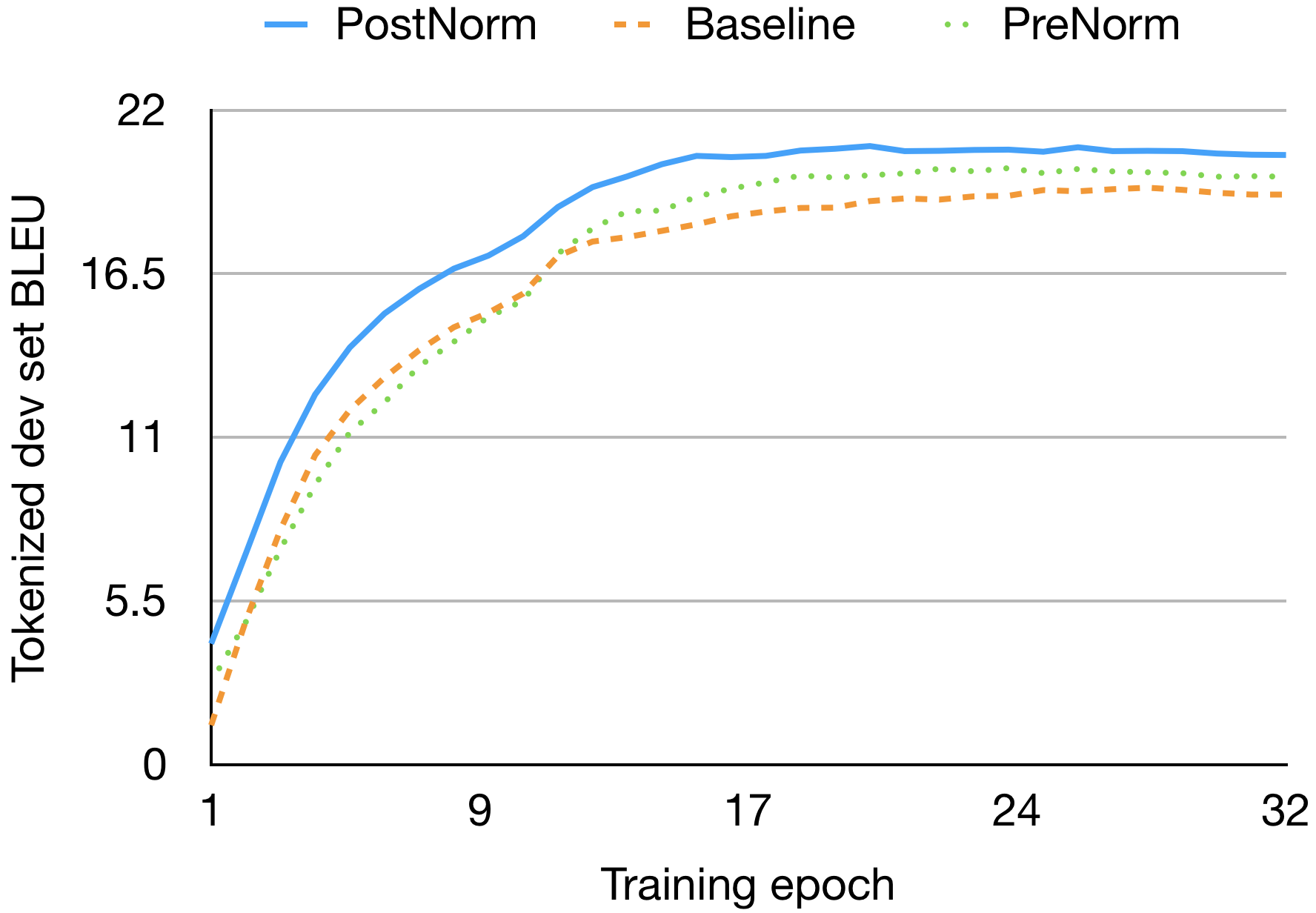}
\caption{Convergence of NMT training with and without LM on English-Turkish.}\label{fig:training}
\end{figure}

\paragraph{Training convergence}

We have found that training converges faster under the {\sc PostNorm} loss. Fig.~\ref{fig:training} plots the training curves of our systems. The baseline (orange curve) reaches its maximum of 19.39 BLEU after 28 training epochs. {\sc PostNorm} surpasses this BLEU score already after 12 epochs.

\paragraph{Language model type}

So far we have used recurrent neural network language models~\citep[RNN-LM]{rnnlm} with LSTM cells in all our experiments. We can also parameterize an $n$-gram language model with a feedforward neural network~\citep[FFN-LM]{ffnlm}. In order to compare both language model types we trained a 4-gram feedforward LM with two 512-dimensional hidden layers and 256-dimensional embeddings on Turkish monolingual data. Tab.~\ref{tab:rnn-vs-ngram-results} shows that the {\sc PreNorm} strategy works particularly well for the $n$-gram LM. However, using an RNN-LM with the {\sc PostNorm} strategy still gives the best overall performance. Using both RNN and $n$-gram LM at the same time does not improve translation quality any further (Tab.~\ref{tab:logprob-results}).

\begin{table}[t!]
\centering
\small
\begin{tabular}{|l@{\hspace{0.2em}}|@{\hspace{0.4em}}r@{\hspace{0.4em}}|@{\hspace{0.4em}}r@{\hspace{0.4em}}|r|r|}
\multicolumn{5}{c}{{\bf English-Turkish (WMT, single system)}} \\\hline
{\bf Method} & \multicolumn{2}{c|}{{\bf Dev set}} & \multicolumn{2}{c|}{{\bf Test set}} \\
 & {\bf FFN} & {\bf RNN} & {\bf FFN} & {\bf RNN} \\\hline
Baseline (no LM) & \multicolumn{2}{c|}{12.23} & \multicolumn{2}{c|}{11.56} \\
Shallow fusion & 12.25 & 12.45 & 11.53 & 11.61 \\
Cold fusion & 12.33 & 12.39 & 11.51 & 11.54 \\\hline
{\bf This work}: {\sc PreNorm} & 12.76 & 12.82 & 11.82 & 11.93 \\
{\bf This work}: {\sc PostNorm} & 12.65 & 13.30 & 11.79 & 12.27 \\
\hline
\end{tabular}

\caption{Comparison between using a recurrent LM (RNN) and an $n$-gram based feedforward LM (FFN) on English-Turkish.}\label{tab:rnn-vs-ngram-results}
\end{table}

\begin{table}[t!]
\centering
\small
\begin{tabular}{|c|c||r|r|r|r|}
\multicolumn{6}{c}{{\bf English-Turkish (WMT), {\sc PostNorm} strategy}} \\\hline
\multicolumn{2}{|c||}{{\bf LM type}} & \multicolumn{2}{c|}{{\bf Single}} & \multicolumn{2}{c|}{{\bf 4-Ensemble}} \\
 {\bf FFN} & {\bf RNN} & {\bf dev} & {\bf test} & {\bf dev} & {\bf test} \\\hline
 & & 12.23 & 11.56 & 14.17 & 13.35 \\
$\checkmark$ & & 12.65 & 11.79 & 14.36 & 13.48 \\
 & $\checkmark$ & 13.30 & 12.27 & 14.77 & 13.61 \\
$\checkmark$ & $\checkmark$ & 12.86 & 12.02 & 14.72 & 13.70 \\
\hline
\end{tabular}
\caption{Combining an RNN-LM and a feedforward LM with the translation model using the {\sc PostNorm} strategy.}\label{tab:logprob-results}
\end{table}

\begin{table}[b!]
\centering
\small
\begin{tabular}{|l|r|r|}
\hline
 {\bf Method} & {\bf Perplexity} & {\bf Average entropy}  \\\hline
 Baseline (no LM) & 23.46 & 3.19 \\
 RNN-LM & 59.19 & 4.66 \\
 TM under {\sc PostNorm} & 113.69 & 1.82 \\
\hline
\end{tabular}
\caption{Perplexity and average entropies of the distributions generated by our systems on the English-Turkish dev set.}\label{tab:entropy}
\end{table}

\begin{table*}[t!]
\centering
\small
\begin{tabular}{|l||r||r|r|r|r||r|}
\hline
 {\bf Method} & {\bf BLEU} & \multicolumn{4}{c||}{{\bf Precisions}} & {\bf BP} \\
 & & {\bf 1-gram} & {\bf 2-gram} & {\bf 3-gram} & {\bf 4-gram} &  \\\hline
 Baseline (no LM) & 17.91 & 53.0 & 23.7 & 12.3 & 6.6 & 0.996\\
{\bf This work}: {\sc PreNorm}  & 19.01 & 54.0 & 24.9 & 13.4 & 7.4 & 1.000 \\\hline
Relative improvement & +6.14\% & {\bf +1.89\%} & {\bf +5.06\%} & {\bf +8.94\%} & {\bf +12.12\%} & -- \\
\hline
\end{tabular}
\caption{BLEU $n$-gram precisions for Estonian-English.}\label{tab:bleu-precisions}
\end{table*}

\begin{table*}[t!]
\centering
\small
\begin{tabular}{|l|l|}
\hline
{\bf Source} & Eestis ja Hispaanias peeti kinni neli Kemerovo grupeeringu liiget \\
{\bf Reference} & Four members of the Kemerovo group arrested in Estonia and Spain \\
{\bf Baseline (no LM)} & In Estonia and Spain, four kemerovo groups were held \\
{\bf This work ({\sc PreNorm})} & Four Kemerovo group members were held in Estonia and Spain \\
\hline
{\bf Source} & Ta ütleb, et elab aastaid hiljem endiselt hirmus.  \\
{\bf Reference} & He says that years later, he still lives in fear. \\
{\bf Baseline (no LM)} & He says that, for years, he still lives in fear. \\
{\bf This work ({\sc PreNorm})} & He says that many years later he still lives in fear. \\
\hline
{\bf Source} & ``Ma kardan," ütleb ta. \\
{\bf Reference} & ``I'm afraid," he says. \\
{\bf Baseline (no LM)} & ``I fear," says he. \\
{\bf This work ({\sc PreNorm})} & ``I am afraid," he says. \\
\hline
\end{tabular}
\caption{Translation samples from the Estonian-English test set.}\label{tab:samples}
\end{table*}

\paragraph{Impact on the TM distribution}

With the {\sc PostNorm} strategy, the TM still produces a distribution over the target vocabulary as the scores are normalized before the combination with the LM. This raises a natural question: How different are the distributions generated by a TM trained under {\sc PostNorm} loss from the distributions of the baseline system without LM? Tab.~\ref{tab:entropy} gives some insight to that question. As expected, the RNN-LM has higher perplexity than the baseline as it is a weaker model of translation. The RNN-LM also has a higher average entropy which indicates that the LM distributions are smoother than those from the baseline translation model. The TM trained under {\sc PostNorm} loss has a much higher perplexity which suggests that it strongly relies on the LM predictions and performs poorly when it is not combined with it. However, the average entropy is much lower (1.82) than both other models, i.e. it produces much sharper distributions.

\paragraph{Language models improve fluency}

A traditional interpretation of the role of an LM in MT is that it is (also) responsible for the fluency of translations~\citep{smt}. Thus, we would expect more fluent translations from our method than from a system without LM. Tab.~\ref{tab:bleu-precisions} breaks down the BLEU score of the baseline and the {\sc PreNorm} ensembles on Estonian-English into $n$-gram precisions. Most of the BLEU gains can be attributed to the increase in precision of higher order $n$-grams, indicating improvements in fluency. Tab.~\ref{tab:samples} shows some examples where our {\sc PreNorm} system produces a more fluent translation than the baseline.

\begin{figure}[t!]
\centering
\includegraphics[width=0.48\textwidth]{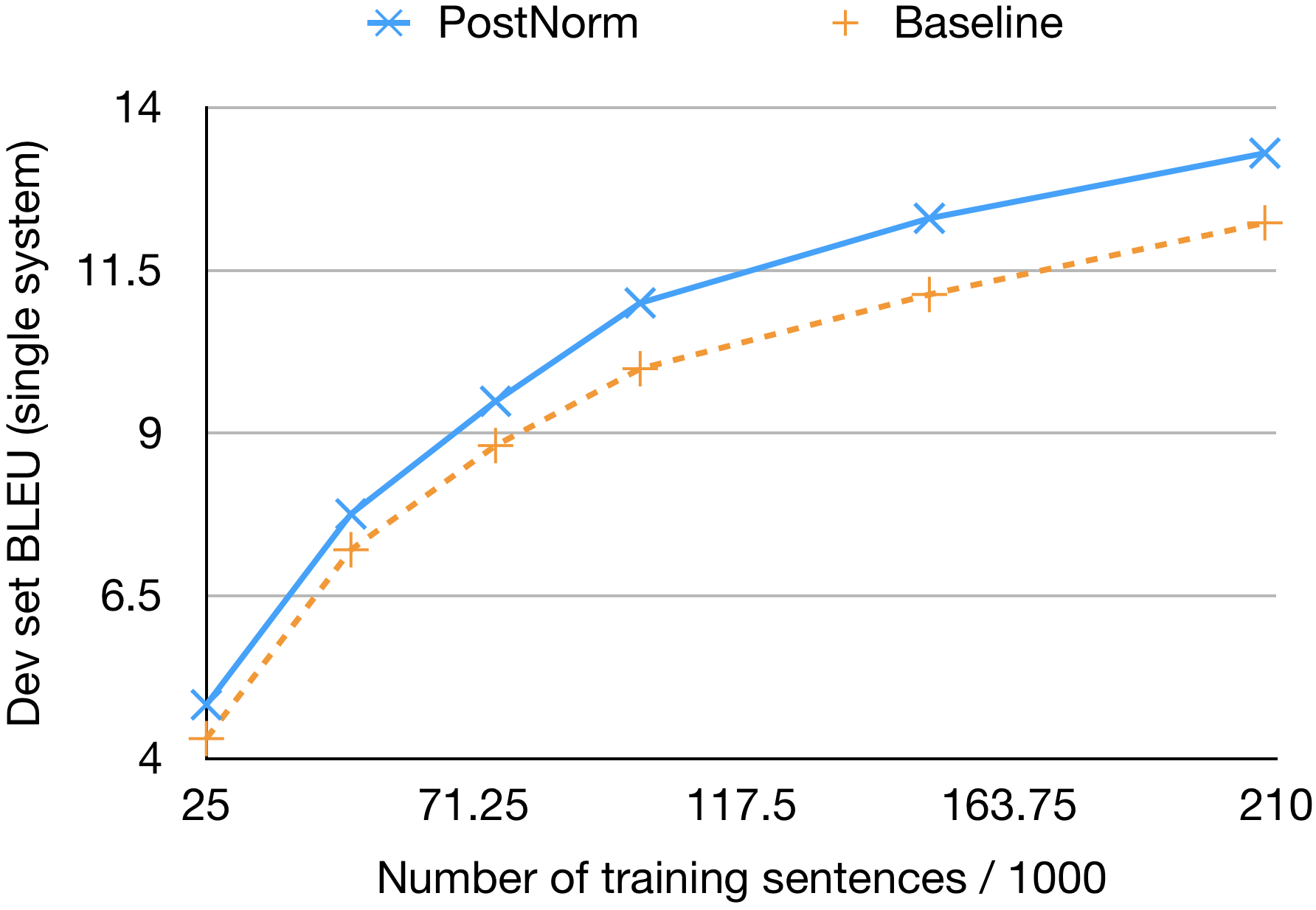}
\caption{English-Turkish BLEU over training set size.}\label{fig:train-set-size}
\end{figure}

\paragraph{Training set size}

We artificially reduced the size of the English-Turkish training set even further to investigate how well our method performs in low-resource settings (Fig.~\ref{fig:train-set-size}). Our {\sc PostNorm} strategy outperforms the baseline regardless of the number of training sentences, but the gains are smaller on very small training sets.

\section{Conclusion}

We have presented a simple yet very effective method to use language models in NMT which incorporates the LM already into NMT training. We reported significant and consistent gains from using our method in four language directions over two alternative ways to integrate LMs into NMT ({\em shallow fusion} and {\em cold fusion}) and showed that our approach works well even in combination with backtranslation and on top of ensembles. Our method leads to faster training convergence and more fluent translations than a baseline system without LM.

\bibliography{refs}
\bibliographystyle{acl_natbib_nourl}


\end{document}